\documentclass{article}

\PassOptionsToPackage{numbers, compress}{natbib}

\usepackage[final, nonatbib]{neurips_2020}

\usepackage[english]{babel}
\usepackage[utf8]{inputenc} 
\usepackage[T1]{fontenc}    
\usepackage{hyperref}       
\usepackage{url}            
\usepackage{booktabs}       
\usepackage{amsfonts}       
\usepackage{nicefrac}       
\usepackage{microtype}      
\usepackage{graphicx}
\usepackage{subcaption}
\usepackage{caption} 
\usepackage{varwidth}
\usepackage{bm}
\usepackage{amsmath}
\title{Crowd-Sourced Road Quality Mapping in the Developing World}

\author{%
  Benjamin Choi \\
  Department of Electrical Engineering\\
  Stanford University\\
  \texttt{benchoi@stanford.edu} \\
  \And
  John Kamalu \\
  Department of Computer Science\\
  Stanford University\\
  \texttt{jkamalu@cs.stanford.edu}
}

\begin{document}

\maketitle

\begin{abstract}

Road networks are among the most essential components of a country’s infrastructure. By facilitating the movement and exchange of goods, people, and ideas, they support economic and cultural activity both within and across borders. Up-to-date mapping of the the geographical distribution of roads and their quality is essential in high-impact applications ranging from land use planning to wilderness conservation. Mapping presents a particularly pressing challenge in developing countries, where documentation is poor and disproportionate amounts of road construction are expected to occur in the coming decades. We present a new crowd-sourced approach capable of assessing road quality and identify key challenges and opportunities in the transferability of deep learning based methods across domains.

\end{abstract}

\section{Introduction}

As the world rapidly urbanizes, road networks follow suit, enabling the transport of essential services and resources to support this development. This growth is expected to occur disproportionately in the developing world, with non-OECD countries expected to account for 90\% of global road infrastructure development over the next 40 years and African "development corridors" expected to add upwards of 53,000 km of new roads \cite{dulac2013global, Laurance2015}. Roads associated with rapid development are often poorly documented or illegal--there are roughly 3 km of illegal or unofficial roads for every 1 km of legal roads in the Brazilian Amazon \cite{alamgir2017economic}.

Road quality is a commonly used proxy for tracking economic activity and development \cite{fan2005road, gertler2016road}. Policymakers and conservationists therefore often rely on up-to-date maps of road quality to inform where resources can most effectively be applied. Recent advances in deep learning have adapted models originally developed for biomedical image segmentation to achieve strong performance in automatic road extraction (i.e. segmentation) tasks from satellite imagery \cite{DBLP:journals/corr/abs-1711-10684}. However, the binarized outputs of these models tend to preserve only structural features, otherwise treating a bustling highway and rural road equivalently. Therefore, road quality classification has been layered on top of segmentation outputs as a more explicit development indicator. Recent road classification methods \cite{Cadamuro:2019:SSM:3314344.3332493, cadamuro2018assigning} have relied on data measured according to the International Roughness Index (IRI), a measurement scheme that requires sophisticated instruments and technical expertise that is difficult and expensive to maintain for a considerable amount of road coverage. Different countries also feature highly varied terrain, raising the possibility of poor performance in out-of-country domains.

We extend current methods in road quality classification to leverage accessible crowdsourced data from OpenStreetMap. Contributors consist of a global set of volunteers and governmental organizations, emphasizing the use of local knowledge and expertise to produce accurate, up-to-date maps \cite{OpenStreetMap}. We use OpenStreetMap to sample satellite imagery along a diverse set of roads of varying quality from multiple developing countries. This imagery is used to train a deep learning road quality classification model. We then apply data masks to highlight the influence of context on model performance. Such methods enable us to anticipate and address the challenges of multi-domain applications of modern classification models. We therefore propose two primary contributions: 1.) the development of a crowd-sourced approach to road quality classification and 2.) a new strategy for diagnosing issues and improving performance in out-of-country domain applications.

\section{Methodology}

\begin{figure}
    \centering
    \includegraphics[width=\linewidth]{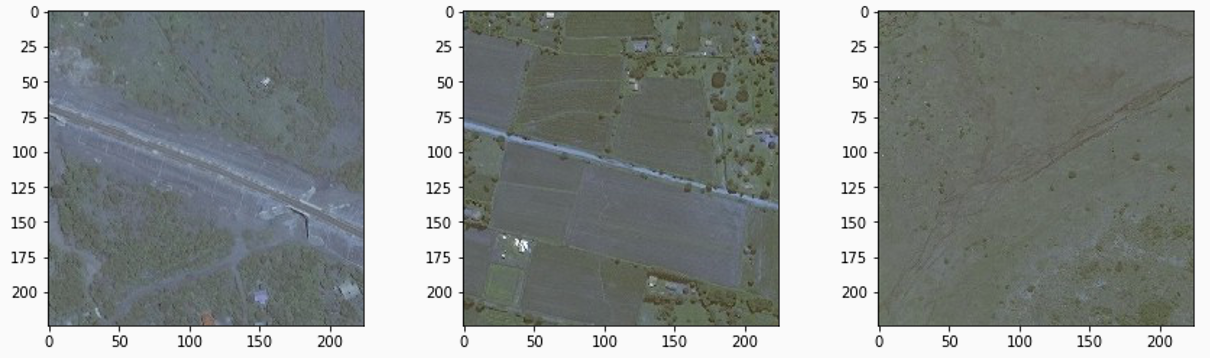}
    \caption{Example RGB images of three road quality classes: major, minor, two-track (left to right).}
    \label{fig:three-class}
\end{figure}

\subsection{Data}

Our input data consists of 30 cm resolution satellite imagery from DigitalGlobe \cite{8575485} combined with road shape information and classification labels derived from OpenStreetMap. In particular, one road-label pair will contain: one $1000px^2$ satellite image covering a $300m^2$ area centered on a given road, a classification of the road belonging to the label set \texttt{\{major, minor, two-track\}}, and GPS coordinates along each road (the polyline). Each classification consists of multiple aggregated subclassifications to increase the resilience of crowd-sourced labels to varying standards of classification from country to country. Figure \ref{fig:three-class} contains representative examples of each road type.

We choose Kenya and Peru as our country domains based on their development status and high potential for future road expansion. Points are sampled from coordinates supplied from OpenStreetMap along each road. Our initial dataset--limited by availability in the OSM database and natural frequency of different road classifications--consists of 170K images in Kenya with a 1:20:12 \textbf{\texttt{major}:\texttt{minor}:\texttt{two-track}} ratio and approximately 97K images in Peru with a 1.3:2:1 ratio. We also enforce a simple threshold on RGB band values to remove images occluded by clouds. To mitigate class imbalances, for each country we randomly sample 15K images balanced evenly across all classifications, and then split this balanced subset with a train:test split of 9:1. The balanced subset is used for all subsequent experiments detailed in this paper.

\subsection{Road Quality Classifier}

To speed up training, we use pre-trained ImageNet \cite{imagenet_cvpr09} classifiers as a foundational model for transfer learning. We append a feed-forward network to these models to translate the output embeddings of "headless" models into a final classification of road quality. Models are evaluated with classification accuracy and F1 score using the randomly sampled test set composed of the 10\% (roughly 1.5K) of images not used in training.

A standard ResNet model is used as a baseline classifier \cite{kaiming2016resnetv2}. For all models, we optimize using categorical cross-entropy loss and an Adam optimizer with a learning rate of 0.0001. After fine-tuning our baseline ResNet model on de-clouded data (using only Kenyan images, for speed of development), we try other CNN architectures and find that in practice, Xception \cite{DBLP:journals/corr/Chollet16a}, a model based on depthwise separable convolutions, achieves strong performance. Therefore, all non-baseline experiments use the Xception model. These experiments also train and test on images from the same country.

Finally, we evaluate the generalizability of our model using an out-of-country domain. Therefore, in contrast to the previous experiments, we use Kenya images for training and Peru images for testing, and vice versa.

\subsection{Strategies for Diagnosing Domain Transferability}

A serious limitation of modern road classification methods is generalizability to disparate domains (i.e. different countries or regions), as expensive recollection of data and retraining therefore becomes necessary to accommodate the domain shift. We attempt to identify and mitigate the primary challenges in out-of-country domain performance. We note that our images consist of both road surfaces and surrounding terrain. We therefore investigate the effect of each image component in isolation. In particular, we are interested in assessing the extent to which modern road classification methods rely on surrounding geographical context for classification, as highly variable geography found across the globe could then drastically affect model performance.

Diagnosis of dependence on geographical context in road classification is conducted using a simple binary mask generated by the following procedure. For each satellite image, geographic points associated with the road line geometry are sampled. These points are connected to produce a road line, with the final pixel coordinates of the road generated using the Bresenham line algorithm \cite{Bresenham:1965:ACC:1663347.1663349}. Each mask pixel is then expanded radially to fully cover the area of the image representing roads. Pixel-wise multiplication of our raw satellite image with our generated mask therefore results in context-occluded images. We also invert the road mask to train only-context models for comparison. The performance of our model in differing masking scenarios acts as a proxy for the extent to which our models utilize actual road pixels in classification of the overall image. We use a balanced dataset for training to eliminate unwanted effects from class imbalances.

\section{Results}

\begin{figure}[]
\begin{minipage}{0.45\linewidth}
    \centering
    \begin{tabular}{ l r r r }
      \toprule
      & \textbf{Accuracy} & \textbf{F1} \\
      \midrule
      \textbf{Peru - No Mask} & 67\% & 66\% \\
      \textbf{Kenya - No Mask} & 80\% & 73\% \\
      \textbf{Kenya - Context Occluded } & 77\% & 72\% \\
      \textbf{Kenya - Road-Occluded} & 79\% & 71\% \\
      \bottomrule
    \end{tabular}
    \captionsetup{justification=centering}
    \captionof{table}{Results from masking experiments. Accuracies reported here are unweighted.}
    \label{table:xception}
    \begin{tabular}{ l r r r }
      \toprule
      \textbf{Train}& \textbf{Test} & \textbf{Accuracy} \\
      \midrule
      Kenya & Peru & 46\% \\
      Peru & Kenya & 60\% \\
      \bottomrule
    \end{tabular}
    \captionof{table}{Results from out-of-country domain experiments.}
    \label{table:domain}
\end{minipage}\hfill
\begin{minipage}{0.45\linewidth}
\centering
    \begin{subfigure}[b]{0.47\linewidth}
    \includegraphics[width=\linewidth]{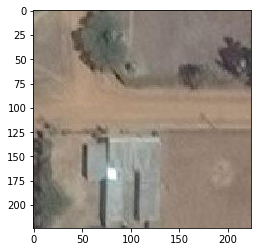}
    \end{subfigure}
    \begin{subfigure}[b]{0.47\linewidth}
    \includegraphics[width=\linewidth]{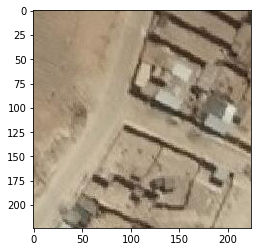}
    \end{subfigure}
    \caption{A Kenyan minor road classified as a two-track road by Peru-to-Kenya model (left) and a true Peruvian two-track road (right). Their similarity conveys the difficulty in accommodating domain shifts between regions.}
    \label{fig:difficult}
\end{minipage}
\end{figure}

\textbf{Road Quality Classifier:} Our Xception model achieves 80\% classification accuracy on Kenyan images, a substantial improvement over the 70\% classification accuracy achieved by the baseline ResNet model. In Table \ref{table:xception}, we find that a pre-trained Xception model outperforms this baseline considerably in the \textit{no mask} scenario. Relatively strong performance across road types and countries suggests that the use of crowd-sourced labels may be a vastly more scalable alternative to IRI data collection. 

Based on Table~\ref{table:domain}, we find that there is a significant decrease in model performance when applying Kenya-trained models to Peru, while there is a smaller decrease when applying Peru-trained models to Kenya. The variability of performance based on direction of transfer suggests that some countries may have country-specific signals (e.g. Kenya) relevant to classification, while others may have more general signals (e.g. Peru), possibly allowing for training models more ideally-equipped for robust classification across domains. Our results indicate that implementation at scale will require relatively granular, localized training data specific to regions of interest in order to accurately capture associations with context. In contrast to traditional IRI measurements, crowd-sourced measurements can be captured and shared by anyone at a hyperlocal resolution. The strength of crowd-sourced methods is therefore emphasized by cross-domain experiments.

While simply masking context may seem like a trivial solution to avoiding reliance on country-specific context, this is undesirable for several reasons. First, the appearance of roads in different classifications are country-specific, as overhanging vegetation, vehicle types, and even soil color (e.g. for two-track roads) all have strong associations with specific locales. Further, when road quality is used as a proxy for development, the state of the surrounding built environment may be a desired component in the overall model's assessment of the extent of urbanization. The crowd-sourced road quality classification models presented here may act as a benchmark as the use of crowd-sourced data becomes increasingly viable and advantageous over manually collecting data.

\textbf{Strategies for Diagnosing Domain Transferability:} Table~\ref{table:xception} also summarizes results from each masking approach. Both models achieve similar performance to our non-occluded baseline, with road-occluded models performing slightly better than context-occluded models. These results support the notion that road classification models tend to rely heavily on context when available, and are even able to classify “roads” with reasonable performance without the actual road being present in the image. The comparable performance of road-occluded models also suggests that road quality classifications may have strong associations with geographical context.  Our results suggest that there is a strong association between road type and surrounding geographic context. However, our context-occluded models still achieve reasonable performance, suggesting that the road itself does produce a signal relevant to the model.

\begin{figure}
    \centering
    \includegraphics[width=\linewidth]{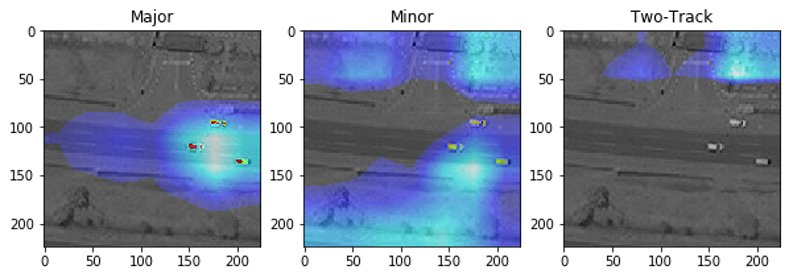}
    \caption{Class activation maps generated for a Kenyan major road reveal disparate activation regions.}
    \label{fig:cam}
\end{figure}

The relative invariance of our model to occlusion of context and road can be better understood through the use of class activation maps \cite{DBLP:journals/corr/ZhouKLOT15}. As shown in Figure \ref{fig:cam}, class activation maps highlight per-class discriminative image regions. While major roads do seem to focus on relevant road areas of the image, minor and two-track roads often seem to rely on contextual information for classification. The occlusion invariance of our model may therefore be attributed to differences in model attention regions across classes.

The results presented here do not necessarily imply that extremely localized image-label pairs are necessary to achieve a model that can accurately associate the unique context of a region with road types. While greater investment in crowd-sourcing up-to-date labels could be advantageous, at some intermediate amount of crowd-sourced coverage within a country, models should be able to generalize effectively to unseen images.

\section{Conclusion and Future Work}

We have shown that models that allow for the use of surrounding context in road classification may be more susceptible to failure cases in applications to out-of-country or out-of-region applications. Surrounding vegetation, land use, building styles, and other contextual factors likely possess at least a weak association with the specific country or region of interest. Careful curation of datasets used for training road and infrastructure classification models is consequently essential. Depending on the application, sampling of datasets should occur either within close local proximity of the region of interest, or with diverse multi-region input for more generalized applications. It is important to note that the adequacy of crowd-sourced labels as a proxy for in-situ road quality measurements such as IRI has not been formally verified beyond visual inspection. Crowd-sourced labels are also likely to suffer from more noise than a single expert annotator. However, in rapidly developing regions with no other road quality measurement system, our methods represent a useful first pass in tracking global road quality.

The development of methods that leverage crowd-sourcing is crucial in ensuring equitable, democratized use of data that can strongly influence the course of events in some of the most rapidly developing regions of the world. We have seen that models struggle to generalize to multiple domains, necessitating the collection of localized data. Producing measurements like the IRI at scale is slow, resource-intensive, and possibly overkill, further encouraging the use of crowd-sourcing labels as an alternative. As internet access becomes more widespread, the power of local infrastructure development tracking may be shifted through crowd-sourcing from governmental and scientific institutions directly to people living through this change.

Future work may address the temporal dimension of road quality tracking. While we have highlighted the challenges posed by spatial variability in the image domain, it is also possible that anthropogenic climate change and other ecologically-influential forces may exert strong shifts in the image domain over time. With the addition of the temporal dimension it may be possible to not only track development in real-time, but to predict its future course.

\small

\bibliographystyle{unsrt}
\bibliography{nips}

\end{document}